\documentclass[11pt]{article}
\usepackage{acl2015}
\usepackage{times}
\usepackage{url}
\usepackage{latexsym}

\usepackage{amssymb}
\usepackage{amsmath}
\NewDocumentEnvironment{alignb}{b}{%
  \begin{align*}
  \refstepcounter{equation} #1 \tag{\theequation}
  \end{align*}
}{\ignorespacesafterend}
\allowdisplaybreaks
\usepackage{makecell}
\usepackage{multirow}
\usepackage{longtable}
\AtBeginEnvironment{longtable}{\tiny}
\usepackage[font=normalsize]{caption} 
\usepackage{graphicx}
\usepackage{subfloat}
\usepackage{subcaption}
\usepackage[numbers]{natbib}
\usepackage{listings}
\usepackage{color}
\usepackage{bbm}
\usepackage{amsfonts}
\usepackage{esvect}
\usepackage{tabularx}

\title{Revolutionizing Brain Tumor Imaging: Generating Synthetic 3D FA Maps from T1-Weighted MRI using CycleGAN Models}
{\footnotesize
\author{XIN DU \\  University of Cambridge\\
  {\tt xd260@cam.ac.uk} \\\And
  Francesca M. Cozzi\\
  Cambridge Brain\\Tumour Imaging Lab\\
  University of\\ Cambridge\\
  {\tt fmg39@cam.ac.uk}\\\And
  Rajesh Jena \\
  University of Cambridge\\
  {\tt rjena@nhs.net} \\}

\date{}

\begin{document}
\maketitle
\begin{abstract}
Fractional anisotropy (FA) and directionally encoded colour (DEC) maps are essential for evaluating white matter integrity and structural connectivity in neuroimaging. However, the spatial misalignment between FA maps and tractography atlases hinders their effective integration into predictive models. To address this issue, we propose a CycleGAN-based approach for generating FA and DEC maps directly from T1-weighted MRI scans, representing the first application of this technique to both healthy and tumor-affected tissues. Our model, trained on unpaired data, produces high-fidelity maps, which have been rigorously evaluated using Structural Similarity Index (SSIM) and Peak Signal-to-Noise Ratio (PSNR), demonstrating particularly robust performance in tumour regions. Radiological assessments further underscore the model’s potential to enhance clinical workflows by providing an AI-driven alternative that reduces the necessity for additional scans.
\end{abstract}

\section{Introduction}
Fractional anisotropy (FA) maps are crucial for assessing white matter integrity by quantifying the directional diffusivity of water molecules in brain tissue. These maps provide valuable insights into brain microstructure and play a fundamental role in neuroimaging, particularly in structural connectivity analysis and tractography-based models \cite{wang2024white,feng2025microstructural,basser2000vivo,le2015diffusion}. Despite their clinical and research significance, the integration of FA maps into analytical pipelines is often hindered by spatial misalignment between standard FA maps and tractography atlases, leading to challenges in cross-modal data fusion and limiting their broader applicability \cite{van2007surface,jones2013white}.

To address this limitation, generating FA maps directly from T1-weighted (T1WI) images presents a promising alternative. T1WI scans are widely available in routine clinical practice and offer high-resolution anatomical details, making them an ideal candidate for synthesizing diffusion-derived metrics. Recent advances in artificial intelligence (AI) have enabled deep learning models to generate high-fidelity medical images, enhancing diagnostic capabilities across multiple modalities, including MRI and CT \cite{karimi2024diffusion}. AI-driven generative models, such as deep learning-based image translation networks, have demonstrated the ability to synthesize missing imaging modalities with high accuracy, improving image quality while reducing dependency on specialized sequences \cite{preetha2021deep,herrmann2024shortening,abu2021paired}. These advancements are particularly beneficial for patients who may not have access to diffusion-weighted imaging (DWI) due to scanning constraints or contraindications, such as renal insufficiency or sensitivity to contrast agents.

In this study, we propose using Cycle Generative Adversarial Network (CycleGAN) framework \cite{zhu2017unpaired} to generate 3D FA and directionally encoded colour (DEC) maps directly from T1WI scans for both healthy brains and tumour-affected brains. Unlike traditional deep learning models that require perfectly paired training data, CycleGAN leverages unpaired image-to-image translation, making it particularly suitable for medical imaging applications where annotated datasets are often scarce and different modalities are hard to register perfectly. By synthesizing FA maps from T1WI directly, our approach ensures spatial consistency with atlases, facilitating seamless integration into predictive models while enhancing the diagnostic utility of conventional MRI scans.

This study aims to establish the feasibility of AI-driven 3D FA synthesis, demonstrating its capability to preserve high-fidelity representations of white matter structure and pathological regions, such as tumours. The integration of this approach into clinical workflows has the potential to expand access to diffusion-derived biomarkers, improve neuroimaging analyses, and enhance patient care by reducing the need for additional imaging sequences.

\section{Related work}
The application of advanced machine learning models for cross-modality generation in medical imaging has gained significant attention \cite{dayarathna2024deep}. CycleGAN, emerging as a prominent approach introduced by \citet{zhu2017unpaired}, allows for the learning of mappings between unpaired medical image datasets, making it particularly useful in scenarios where paired training samples are not feasible. Numerous studies have highlighted the efficiency of CycleGAN in the transformation of medical images between different modalities \cite{hu2024synthetic, diniz2024cross}.

For example, \citet{gong2024channel} exploited CycleGAN enhanced with channel-wise attention mechanisms to better focus on relevant features within MRI data while simultaneously constraining the structural similarity between synthetic and target CT images. This dual approach effectively integrates the rich contextual information from MRI and ensures that the generated CT images retain high fidelity. This approach minimizes the need for CT scans, thus reducing patient exposure to radiation and the burden associated with multiple imaging sessions and enhancing treatment planning in head and neck radiotherapy. Similarly, Wolterink et al. \cite{wolterink2017deep} applied CycleGAN to synthesize MRI images from CT scans, offering supplemental diagnostic information while reducing overall scanning time and resource utilization \cite{wang2023dc}.

In oncology, \citet{farshchitabrizi2025ai} utilized CycleGAN to explore the application of CycleGAN for synthesizing positron emission tomography (PET) images based on attenuation correction without additional CT imaging, enhancing early cancer screening and monitoring while reducing exposure to harmful radiotracers. This application exemplifies how cross-modality generation can significantly impact patient safety and clinical workflow.

Furthermore, Gu et al. \cite{gu2019generating} explored the application of CycleGAN in estimating FA values from T1WIs using a 2D framework, illustrating the potential of these approaches to provide reliable assessments of white matter integrity without the necessity of diffusion-weighted imaging. These studies collectively demonstrate the transformative impact of CycleGAN in the medical imaging domain, paving the way for innovative diagnostic tools and efficient clinical workflows.  

Beyond CycleGAN, style transfer models have demonstrated potential in image generation. The method by \citet{huang2017arbitrary} uses style representations to combine content from one modality with the style of another. Similarly, \citet{cao2023deep} introduces StyleMapper, a medical image style transfer method that efficiently transforms breast MRI scans into unseen styles using limited training data. By utilizing a varied set of simulated medical imaging styles, StyleMapper enhances computational efficiency and enables arbitrary style transfer, which is crucial for managing different imaging protocols and scanner models in medical imaging. This method helps create a unified style dataset of medical images, improving training for downstream tasks like classification and object detection.

Diffusion models have recently emerged as a powerful alternative for generative tasks, including image synthesis. These models, which work by gradual denoising a random signal to create coherent images, have shown capability in generating high-quality medical images \cite{kazerouni2023diffusion,dhariwal2021diffusion}. \citet{kim2024adaptive} introduces a model based on the latent diffusion model (LDM) for image-to-image translation in 3D medical images, overcoming challenges in acquiring multi-modal images due to cost and safety concerns. Utilizing a switchable block architecture, the model generates high-quality 3D images without patch cropping and addresses limitations of 2D methods. The multiple switchable spatially adaptive normalization (MS-SPADE) block enables various translation tasks with a single model.

These advancements underscore the versatility of various model architectures in tackling challenges in medical imaging, particularly in enhancing accessibility and alleviating patient burden. However, there is a scarcity of studies focused on generating 3D FA maps from normal MRI scans for both healthy and tumour-affected brains, overcoming the shortcoming of the missing out-of-slice information in 2D generation methods. Generating 3D FA maps is crucial for subsequent clinical research due to the complexities involved in spatial alignment and registration of FA maps to template scans. Our work leverages the capabilities of CycleGAN to address this challenge, offering a solution that minimizes the need for additional scan processing beyond routine procedures, thereby reducing the burden on patients.
\begin{figure*}[!h]
    \centering
    \includegraphics[width=\textwidth]{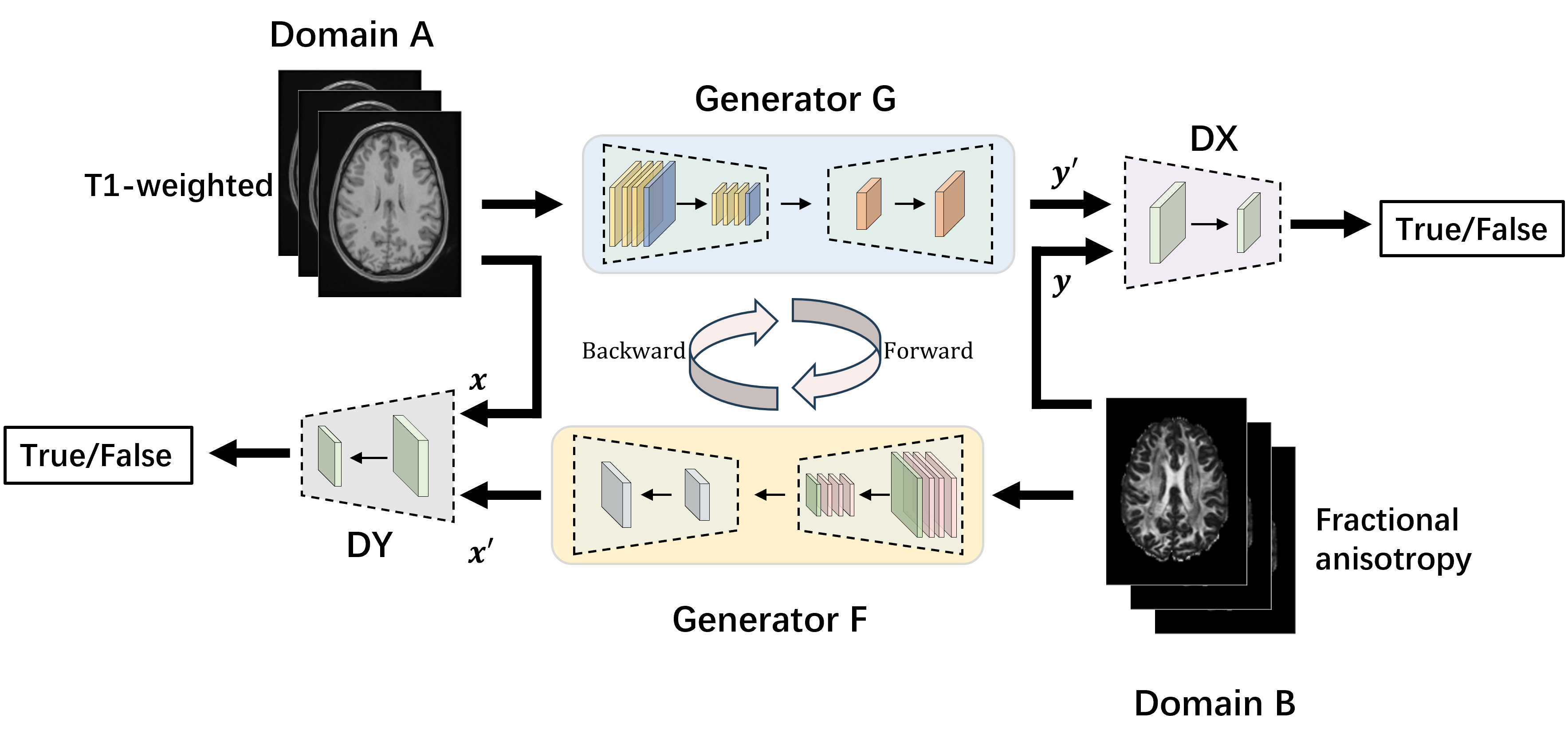}
    \caption{The schematic of the 3D CycleGAN architecture.}
    \label{fig:architecture}
\end{figure*}

\section{Datasets}
Two datasets are used for synthetic image generation. For healthy cases, T1WI are sourced from the Human Connectome Project (HCP)\footnote{Data for this project were collected and shared through the Human Connectome Project (HCP; U01-MH93765), led by Principal Investigators Bruce Rosen, M.D., Ph.D., Arthur W. Toga, Ph.D., and Van J. Weeden, M.D. The HCP was funded by the National Institute of Dental and Craniofacial Research (NIDCR), the National Institute of Mental Health (NIMH), and the National Institute of Neurological Disorders and Stroke (NINDS). Data dissemination is managed by the Laboratory of Neuro Imaging at the University of Southern California.} \cite{glasser2013minimal,van2013wu}, comprising 1,065 subjects. Following the settings of \cite{gu2019generating}, 1,000 randomly selected subjects are used for model training, while the remaining 65 serve as unseen test cases. The ground truth FA maps are obtained via diffusion tensor fitting and MAP-MRI fitting, as introduced by \citet{gu2019generating}.  

\section{Methodology}
\subsection{CycleGAN Architecture}
Cycle-Consistent Generative Adversarial Networks (CycleGAN)~\cite{zhu2017unpaired} facilitate unpaired image-to-image translation by leveraging adversarial training and cycle consistency. The architecture consists of two generators, $G: X \rightarrow Y$ and $F: Y \rightarrow X$, along with two discriminators, $D_X$ and $D_Y$, which distinguish between real and generated images in each domain, as shown in Figure \ref{fig:architecture}.

Following the network architecture design of CycleGAN~\cite{gu2019generating, zhu2017unpaired, ge2019unpaired, ge2019unpaired1, johnson2016perceptual}, a 3D cross-modality medical image generation model is constructed. Specifically, each generator consists of three convolutional layers ($128 \times 128$), nine residual blocks  ($256 \times 256$), and two fractionally-strided convolutional layers. Similar to the configuration proposed by ~\citet{johnson2016perceptual}, instance normalization is employed. Each discriminator is composed of five convolutional layers with LeakyReLU activation and instance normalization. For generating FA maps, both the input and output channels are set to one, whereas for DEC map generation, three channels are used.
\subsection{Adversarial Loss}
To ensure the generated images are indistinguishable from real images, CycleGAN employs adversarial loss for both mappings. For the $X \rightarrow Y$ translation, the objective function is defined as:
\begin{alignb}
\mathcal{L}_{GAN}(G, D_Y, X, Y) &= \mathbb{E}_{y}[\log D_Y(y)] 
                               \\&+ \mathbb{E}_{x}[\log (1 - D_Y(G(x)))]
\end{alignb}
A similar adversarial loss is applied for the $Y \rightarrow X$ mapping.

\subsection{Cycle Consistency Loss}
Since unpaired training data lacks direct supervision, cycle consistency is enforced to ensure $G(F(y)) \approx y$ and $F(G(x)) \approx x$. The cycle consistency loss is formulated as:
\begin{alignb}
\mathcal{L}_{cycle}(G, F) &= \mathbb{E}_{x}[\| F(G(x)) - x \|_1] 
\\&+ \mathbb{E}_{y}[\| G(F(y)) - y \|_1]
\end{alignb}
This constraint preserves content and structure during domain translation.
\subsection{Correlation Coefficient (Cor-Coe) Loss}
Following the setting given by \citet{ge2019unpaired1}, The Correlation Coefficient (Cor-Coe) Loss is designed to measure the similarity between the generated and real images in terms of pixel or feature correlation. Unlike pixel-wise losses (e.g., L1 or L2 loss), Cor-Coe loss captures structural similarity by maximizing the linear correlation between the two images.
\begin{alignb}
   \mathcal{L}_{\text{Cor-Coe}}(G, F) = \frac{\text{Cov}(G(x), x)}{\sigma_G(x) \sigma_x} + \frac{\text{Cov}(F(y), y)}{\sigma_F(y) \sigma_y}
\end{alignb}

\subsection{Full Objective Function}
The overall objective function combines adversarial loss, cycle consistency loss and cor-coe loss:
\begin{alignb}
\mathcal{L}(G, F, D_X, D_Y) &= \mathcal{L}_{GAN}(G, D_Y, X, Y) \\&+ \mathcal{L}_{GAN}(F, D_X, Y, X) \\&+ \lambda \mathcal{L}_{cycle}(G, F)\\&+\beta\mathcal{L}_{\text{Cor-Coe}}(G, F)
\end{alignb}
where $\lambda$ controls the weight of the cycle consistency loss and $\beta$ controls the weight of cor-coe loss. They are all set to 1.

\subsection{Training Details}
CycleGAN is trained using the Adam optimizer with a learning rate of $0.0002$ and momentum parameters $\beta_1=0.5, \beta_2=0.999$. A batch size of $1$ or $8$ is typically used, depending on memory constraints. The model is trained for $200$ epochs, with the first half using a constant learning rate and the latter half employing a linearly decaying learning rate. To improve training stability, least-squares loss is used for adversarial training instead of binary cross-entropy. All inputs are resized to 128-128-64. 

For healthy brains, all models are trained from scratch with random initialization. For tumour cases, both training from scratch and transfer learning are applied. In transfer learning, the model is first pre-trained on healthy cases and then fine-tuned on tumour cases to leverage prior knowledge from healthy brains and assess the impact of knowledge transfer on the generated images.

\subsection{Evaluation Metrics}  
To assess the quality of the generated images, we employ three widely used image similarity metrics: Structural Similarity Index (SSIM), Multi-Scale Structural Similarity Index (MS-SSIM), and Peak Signal-to-Noise Ratio (PSNR), which collectively capture both pixel-level accuracy (PSNR) and perceptual similarity (SSIM and MS-SSIM). Additionally, two radiologists perform a visual assessment to distinguish between the generated and ground truth maps using a custom-developed tool.

\subsubsection{Structural Similarity Index (SSIM)}  
SSIM is a perceptual metric that quantifies the similarity between two images by considering luminance, contrast, and structural information \cite{wang2004image}. Unlike traditional pixel-wise metrics, SSIM accounts for human visual perception, making it more robust to variations in intensity and contrast. It is computed as:  

\begin{equation}  
SSIM(x, y) = \frac{(2\mu_x \mu_y + C_1)(2\sigma_{xy} + C_2)}{(\mu_x^2 + \mu_y^2 + C_1)(\sigma_x^2 + \sigma_y^2 + C_2)}  
\end{equation}  

where \( \mu_x \) and \( \mu_y \) are the mean intensities of images \( x \) and \( y \), \( \sigma_x^2 \) and \( \sigma_y^2 \) are their variances, and \( \sigma_{xy} \) represents their covariance. Constants \( C_1 \) and \( C_2 \) stabilize the division. SSIM values range from -1 to 1, where 1 indicates perfect similarity.  

\subsubsection{Multi-Scale Structural Similarity Index (MS-SSIM)}  
MS-SSIM extends SSIM by evaluating image similarity at multiple scales, enhancing its ability to capture structural distortions across different resolutions \cite{wang2003multiscale}. Instead of computing SSIM at a single scale, MS-SSIM applies a hierarchical approach by downsampling the image iteratively and computing SSIM at each level. The overall MS-SSIM score is computed as a weighted product of SSIM values across scales:  

\begin{equation}  
MS\text{-}SSIM(x, y) = \prod_{i=1}^{M} [SSIM_i(x, y)]^{\alpha_i}  
\end{equation}  

where \( SSIM_i \) represents the SSIM value at scale \( i \), and \( \alpha_i \) are weighting factors. MS-SSIM provides improved perceptual correlation compared to SSIM, particularly for images with varying spatial frequencies.  

\subsection{Peak Signal-to-Noise Ratio (PSNR)}  
PSNR is a commonly used metric that measures the ratio between the maximum possible signal power and the power of corrupting noise in an image \cite{hore2010image}. It is defined as:  

\begin{equation}  
PSNR = 10 \log_{10} \left( \frac{MAX_I^2}{MSE} \right)  
\end{equation}  

where \( MAX_I \) is the maximum pixel intensity value (e.g., 255 for 8-bit images), and \( MSE \) is the mean squared error between the generated and ground truth images. Higher PSNR values indicate better image quality, with typical values ranging from 20 to 50 dB, depending on image fidelity.

\section{Experiments}
FA maps were generated using 3D CycleGAN models. In this section, we compare the generated maps with ground truth maps and assess the quality of the generation from both quantitative and qualitative perspectives.

\begin{figure}[!h]
    \centering
    \includegraphics[width=0.49\textwidth]{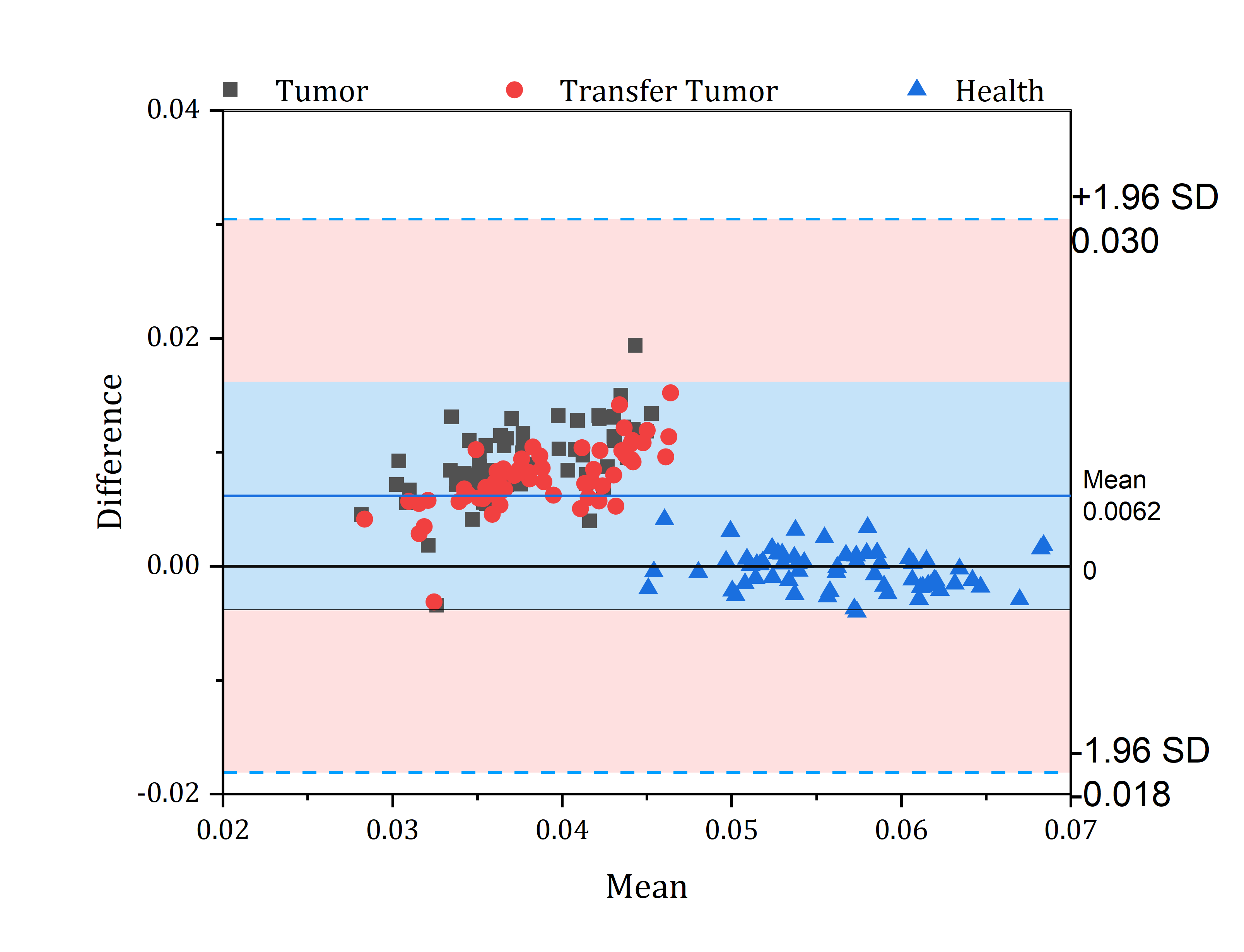}
    \caption{Bland-Altman plot comparing the generated FA maps with the ground truth for both healthy and tumor cases (encompassing results of learning from scratch and transfer learning approaches). }
    \label{fig:stasticofgenerationpixels}
\end{figure} 

\begin{table}[!h]
\centering
\caption{Quantitative evaluation of the generated images using SSIM and MS-SSIM.}
\begin{tabular}{c|cc}
\hline
\bf{Metric}                       & \bf{SSIM}               & \bf{MS-SSIM}            \\ 
\hline
Health FA                   & 0.879 $\pm_{0.024}$    & 0.975 $\pm_{0.024}$    \\ 
Tumor FA                    & 0.883 $\pm_{0.024}$    & 0.958 $\pm_{0.024}$    \\ 
Transfer Tumor FA           & 0.886 $\pm_{0.024}$    & 0.964 $\pm_{0.024}$    \\ 
Health DEC                  & 0.700 $\pm_{0.024}$    & 0.932 $\pm_{0.024}$    \\ 
\hline
\end{tabular}%
\label{tab:ssim-msssim}
\end{table}
\begin{figure}[!h]
    \centering
    \includegraphics[width=0.47\textwidth]{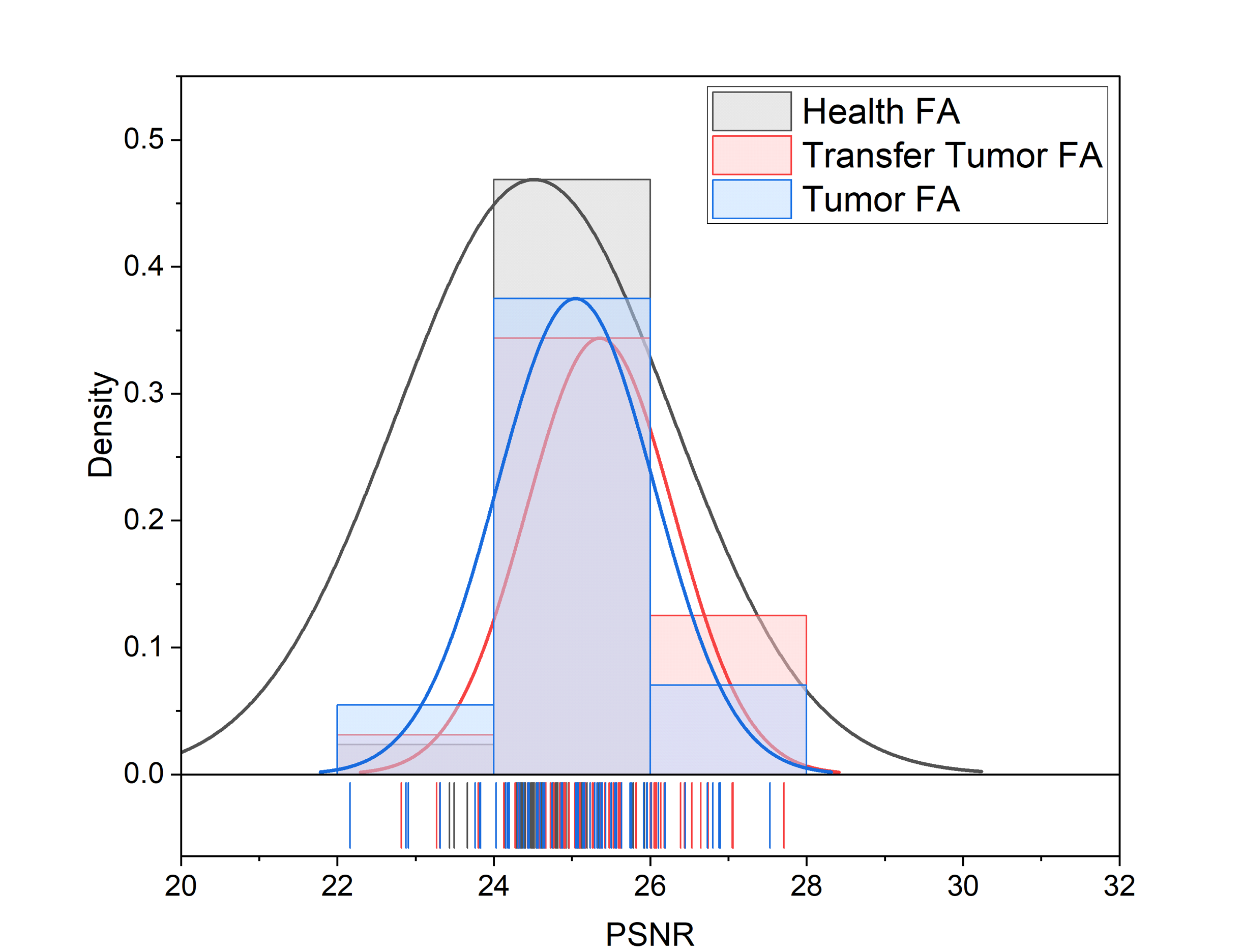}
    \caption{Quantitative evaluation of the generated FA maps using PSNR. Higher PSNR values indicate lower reconstruction error and better image quality.}
    \label{fig:psnrFAmap}
\end{figure}
\begin{figure*}[!h]
    \hfill
    \includegraphics[width=0.8\textwidth]{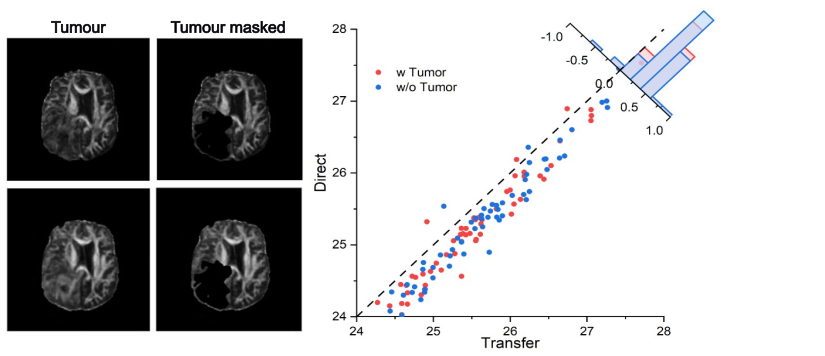}
    \caption{Comparison of generated images with and without tumour masking based on PSNR evaluation. The left subfigure provides an illustration of the tumour masking conditions. The right subfigure presents the PSNR distribution for cases where tumor masking is applied or not, under both transfer learning and direct learning settings. Additionally, the bar plot in the top right corner visualizes the distribution of case-wise PSNR differences between transfer learning and direct learning (\textit{transfer – direct}), providing insights into the impact of transfer learning on reconstruction quality.}
    \label{fig:maskedtumor}
\end{figure*}
\subsection{Quantitative Analysis}
Figure \ref{fig:stasticofgenerationpixels} compares the generated FA maps with the ground truth at the voxel-wise level. The average intensity of each scan is computed for both the generated and ground truth maps for each case. The results show a distinct distribution between tumor and healthy cases. Specifically, the mean intensity value for each 3D scan in healthy cases is approximately 0.055, whereas tumour cases exhibit lower values.

Furthermore, the difference between the model-generated and ground truth maps is statistically lower in healthy cases compared to tumour cases, indicating that the current model performs better in generating FA maps for healthy brains. Additionally, the performance of tumor case generation using transfer learning shows a slightly lower difference compared to training from scratch, suggesting that leveraging knowledge from healthy cases improves the generation of tumour-affected brains. 

Table \ref{tab:ssim-msssim} and Figure \ref{fig:psnrFAmap} present the evaluation of generation quality using SSIM, MS-SSIM, and PSNR. Consistent with the visual assessment, the generated scans for healthy cases demonstrate better performance compared to tumour cases. This discrepancy may be attributed to the higher input quality and a larger number of training samples available for healthy cases.

For the DEC map, the generation performance is notably lower than that of FA maps, indicating greater challenges in accurately synthesizing DEC images. Furthermore, leveraging knowledge transfer enhances the quality of generated tumour case maps, as reflected by an increase in the number of cases with higher PSNR values compared to training without transfer learning.

To further analyze the effects of transfer learning, Figure \ref{fig:maskedtumor} compares the generated images obtained with and without transfer learning. Additionally, it examines the differences between normal tissues (where the tumour is masked) and tumour tissues to assess the preservation of anatomical structures. 

The results indicate that for both the whole brain and the brain regions excluding the tumour, transfer learning slightly outperforms direct learning. This suggests that utilizing knowledge from healthy cases improves the generation quality of both tumour-affected and normal tissue regions, leading to better anatomical consistency in the synthesized images.

\begin{figure*}[!h]
     \begin{subfigure}[b]{0.5\textwidth}
        \centering
    \includegraphics[width=\textwidth]{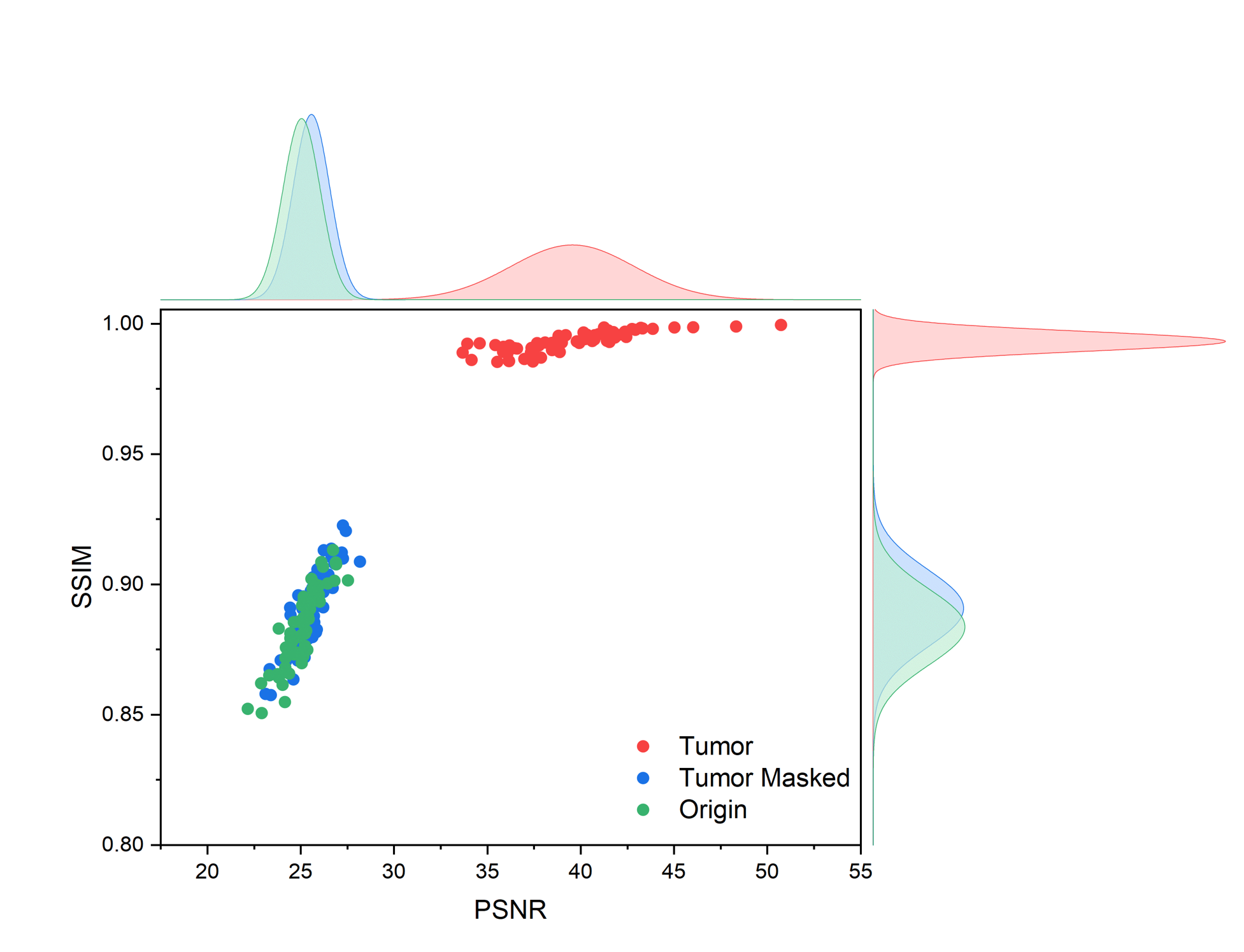}
    \caption{SSIM vs. PSNR}
        \label{fig:ssimpsnr}
     \end{subfigure}
     \hfill
     \begin{subfigure}[b]{0.5\textwidth}
        \centering
    \includegraphics[width=\textwidth]{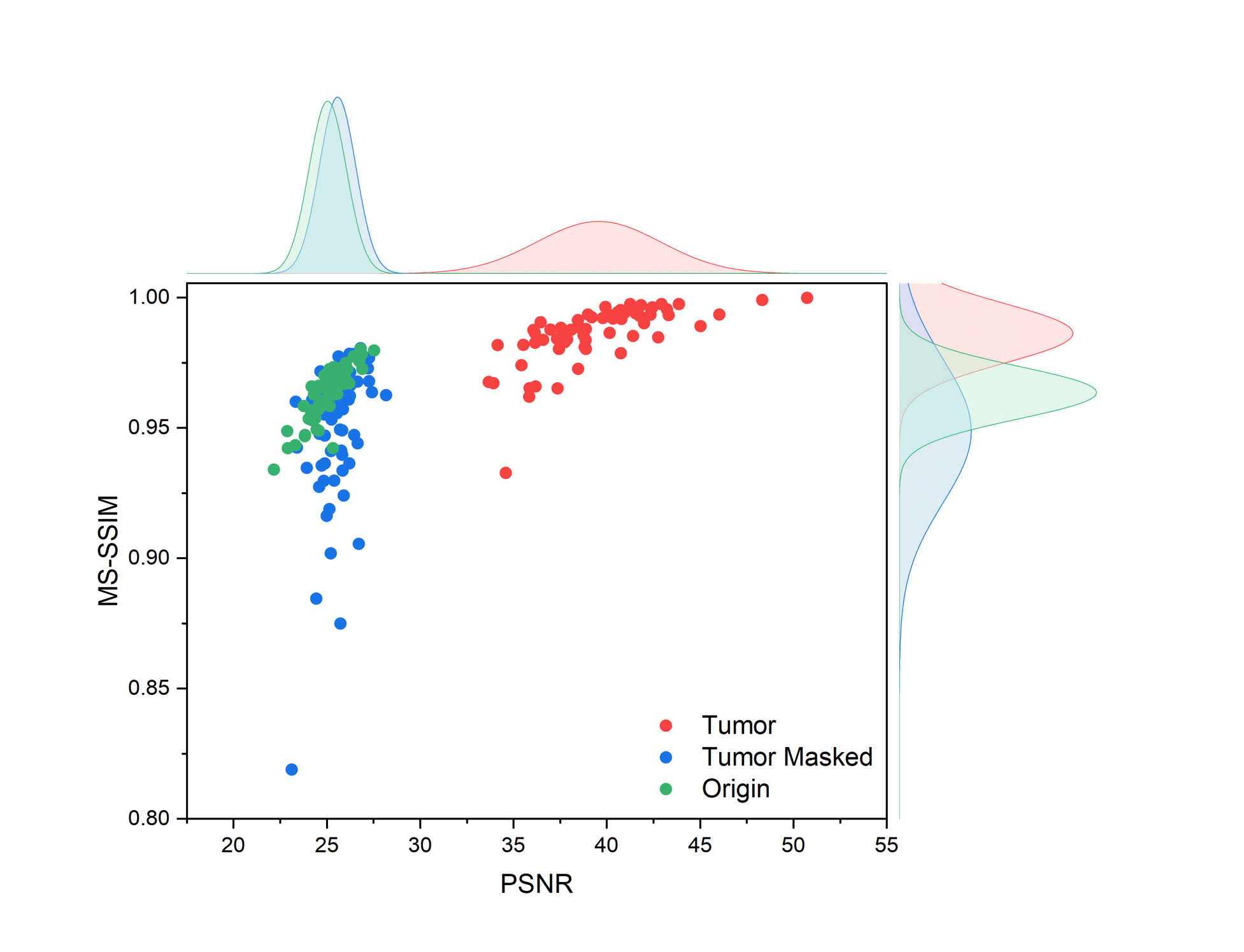}
    \caption{MS-SSIM vs. PSNR}
        \label{fig:massimpasnr}
     \end{subfigure}
\caption{Comparison of generated image quality between the tumour region, the region after tumour masking, and the whole brain image. The evaluation tracks the differences in image quality across these distinct areas.}
\label{fig:evaluationcomparetumormasked}
\end{figure*}

Figure \ref{fig:evaluationcomparetumormasked} provides a comparison of different evaluation metrics to track the correlation between PSNR and either SSIM or MS-SSIM for the tumour region alone, the brain with the tumour masked, and the whole brain. Notably, PSNR shows a stronger positive correlation with SSIM than with MS-SSIM. Additionally, all evaluation metrics demonstrate significantly higher performance on the tumour region alone compared to the brain with the tumour masked or the whole brain. This indicates that the model is better at capturing the details of the tumour during the generation process.

\begin{figure*}[!h]
     \begin{subfigure}[b]{\textwidth}
        \centering
    \includegraphics[width=\textwidth]{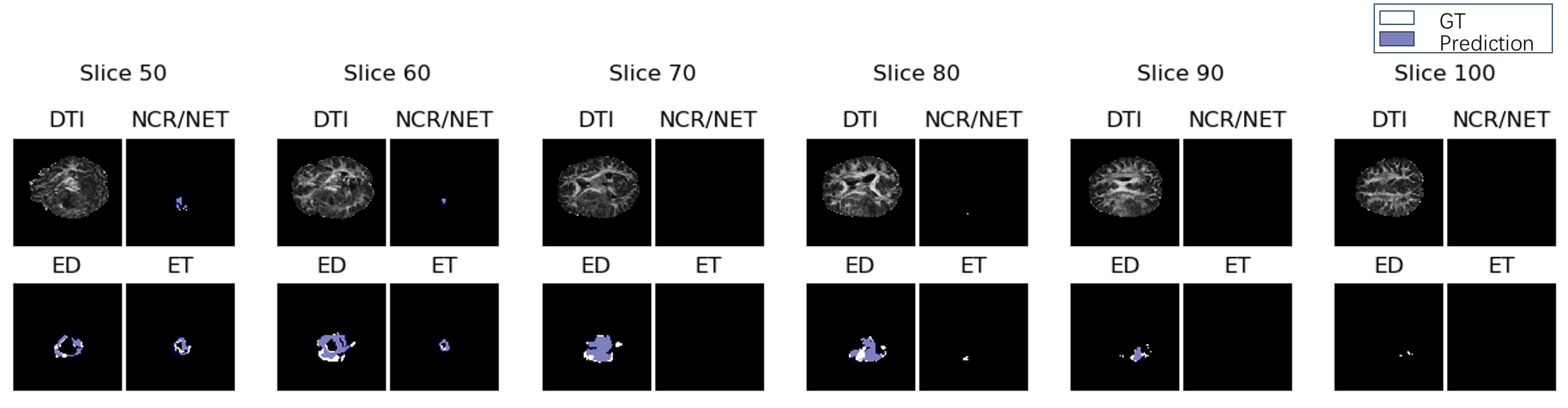}
    \caption{Segmentation based on realistic FA maps}
        \label{fig:segrealFA}
     \end{subfigure}
     \begin{subfigure}[b]{\textwidth}
        \centering
    \includegraphics[width=\textwidth]{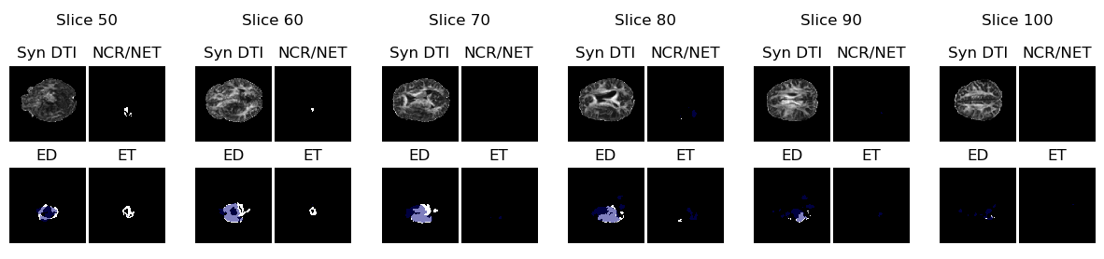}
    \caption{Segmentation comparison of NET/NCR, ED, and ET regions on real and synthetic maps. (\subref{fig:segrealFA}) presents segmentation results on real maps, while (\subref{fig:segsynFA}) displays segmentation results on synthetic maps.}
        \label{fig:segsynFA}
     \end{subfigure}
\caption{Illustration of tumour segmentation.}
\label{fig:segmentation}
\end{figure*}

\subsection{Segmentation}
To further evaluate the quality of the generated maps, a U-Net \cite{isensee2021nnu} is employed for brain tumor segmentation, targeting the non-enhanced tumor (NET) or non-contrast region (NCR), edema (ED), and enhancing tumor (ET) regions. The model is trained on real FA maps and tested on both unseen real FA maps and synthetic FA maps.

Figure \ref{fig:segmentation} presents the segmentation performance on both test sets. The results demonstrate that the model, trained solely on real maps, successfully transfers its segmentation capability to synthetic maps without requiring fine-tuning. Evaluated by the Dice coefficient, the model achieves a score of 0.2793 on synthetic maps (0.1834 for NET, 0.4213 for ED, and 0.2332 for ET) and 0.5695 on real maps (0.5231 for NET, 0.6437 for ED, and 0.5417 for ET). Notably, the segmentation transferability is highest for ED, while performance on NET and ET remains relatively lower. It is important to notice that this segmentation experiment serves as an initial assessment, and improving segmentation performance is not the primary focus of this work. 

\section{Conclusion}
In this study, we investigated the feasibility of generating diffusion tensor imaging (DTI) scans, including fractional anisotropy (FA) and directionally encoded color (DEC) maps, for both healthy and tumor-affected brains using Cycle-GAN models. The quality of the generated maps was assessed through quantitative and qualitative metrics as well as expert evaluation by radiologists.

The results demonstrate that the synthetic maps effectively capture long-range white matter connections, preserving the overall structural integrity of major white matter tracts. However, minor deformations are observed in the representation of local white matter connections, particularly in regions with complex arrangements. These findings suggest that while the proposed approach holds promise for generating high-fidelity DTI data, further refinements are needed to enhance the accuracy of local structural details. Future work will focus on improving the resolution and anatomical precision of the generated maps, ensuring their robustness for clinical and research applications.
\section*{Acknowledgments}
This work was funded in part by a grant from Cancer Research UK, RadNet Cambridge [C17918/A28870], and by the Stella project grant from the International Cancer Expert Corps (ICEC).

\bibliography{reference}
\end{document}